\documentclass[letterpaper, 10 pt, conference]{ieeeconf}  % Comment this line out if you need a4paper

\IEEEoverridecommandlockouts                              % This command is only needed if 
\usepackage[sorting=none]{biblatex}  
\usepackage{graphics} % for pdf, bitmapped graphics files
\usepackage{epsfig} % for postscript graphics files
\usepackage{times} % assumes new font selection scheme installed
\usepackage{amsfonts}
\usepackage{amsmath}

\usepackage{amssymb}
\usepackage[colorlinks, linkcolor=red, citecolor=blue]{hyperref}
\usepackage{algorithm}
\usepackage{algpseudocode}
\usepackage{epsfig}
\usepackage{graphicx}
\usepackage{caption}
\usepackage{subfigure}
\usepackage{mathrsfs}
\usepackage{array}
\usepackage{multirow}
\usepackage{multicol}
\usepackage{booktabs}
\usepackage{makecell}
\usepackage{xcolor}
\usepackage{amsfonts}
\usepackage{gensymb}
\usepackage{flushend}
\usepackage{steinmetz}
\usepackage{threeparttable}

\title{\LARGE \bf
One RING to Rule Them All: Radon Sinogram for \\Place Recognition, Orientation and Translation Estimation
}

\author{Sha Lu$^{1*}$, Xuecheng Xu$^{1*}$, Huan Yin$^{2}$, Zexi Chen$^{1}$, Rong Xiong$^{1}$ and Yue Wang$^{1\dagger}$% <-this % stops a space
\thanks{*This work was supported in part by the National Nature Science Foundation of China under Grant 61903332 and in part by the Zhejiang Provincial Natural Science Foundation of China (LD22E050007).}% <-this % stops a space
\thanks{$^{1}$State Key Laboratory of Industrial Control and Technology and Institute of Cyber-Systems and Control, Zhejiang University, Hangzhou, 310058, China.}
\thanks{$^{2}$Department of Electronic and Computer Engineering, Hong Kong University of Science and Technology, Clear Water Bay, Hong Kong SAR.}%
\thanks{$^{\dagger}$Corresponding author wangyue@iipc.zju.edu.cn.}%
 \thanks{$^{*}$Equal contribution.}%
}

\begin{document}

\maketitle
\thispagestyle{empty}
\pagestyle{empty}

%%%%%%%%%%%%%%%%%%%%%%%%%%%%%%%%%%%%%%%%%%%%%%%%%%%%%%%%%%%%%%%%%%%%%%%%%%%%%%%%
\begin{abstract}
LiDAR-based global localization is a fundamental problem for mobile robots. It consists of two stages, place recognition and pose estimation, which yields the current orientation and translation, using only the current scan as query and a database of map scans. Inspired by the definition of a recognized place, we consider that a good global localization solution should keep the pose estimation accuracy with a lower place density. Following this idea, we propose a novel framework towards sparse place-based global localization, which utilizes a unified and learning-free representation, Radon sinogram (RING), for all sub-tasks. Based on the theoretical derivation, a translation invariant descriptor and an orientation invariant metric are proposed for place recognition, achieving certifiable robustness against arbitrary orientation and large translation between query and map scan. In addition, we also utilize the property of RING to propose a global convergent solver for both orientation and translation estimation, arriving at global localization. Evaluation of the proposed RING based framework validates the feasibility and demonstrates a superior performance even under a lower place density. 
% Appendix is in https://sites.google.com/view/onering.

\end{abstract}

%%%%%%%%%%%%%%%%%%%%%%%%%%%%%%%%%%%%%%%%%%%%%%%%%%%%%%%%%%%%%%%%%%%%%%%%%%%%%%%%
\section{Introduction}
% It is composed of place recognition and pose estimation. Place recognition provides candidates for loop closure detection. Global localization comprises two stages, place recognition and pose estimation.
LiDAR-based global localization is a fundamental problem for mobile robots, which enables the initialization of a navigation system, as well as loop closing in a SLAM system \cite{magnusson2009appearance,dube2020segmap}. The problem is highly challenging since it aims at searching the solution over the entire candidate pose space, i.e., every pose that the robot can move on the map, using only the current scan and a database of map scans. In addition, lightweight computation and storage are also desired due to the constrained resource, thus introducing specific requirements of both effectiveness and efficiency for global localization.
% Besides, scans registration technique like Iterative Closest Point (ICP) is used to estimate the robot's pose. 
 
To address the problem, a great number of works take the strategy originated from visual localization \cite{lowry2015visual,uy2018pointnetvlad,lynen2020large} by dividing the problem into two phases: place recognition first and then pose estimation \cite{kim2018scan,chen2021overlapnet,xu2021disco,ding2022translation}. Place recognition identifies the place where the robot lies, while the pose estimation yields the accurate position and orientation of the robot based on the recognized place. Note that the recognized place serves as an intermediate role between the two phases. From the perspective of candidate pose space discretization, we consider the place as a coarse-level discrete cell, which ensures that the current pose is in this cell. In this context, a question of how to define the discretization resolution of a place naturally arises. 

\begin{figure}[t]
\centering
\includegraphics[width=8.5cm]{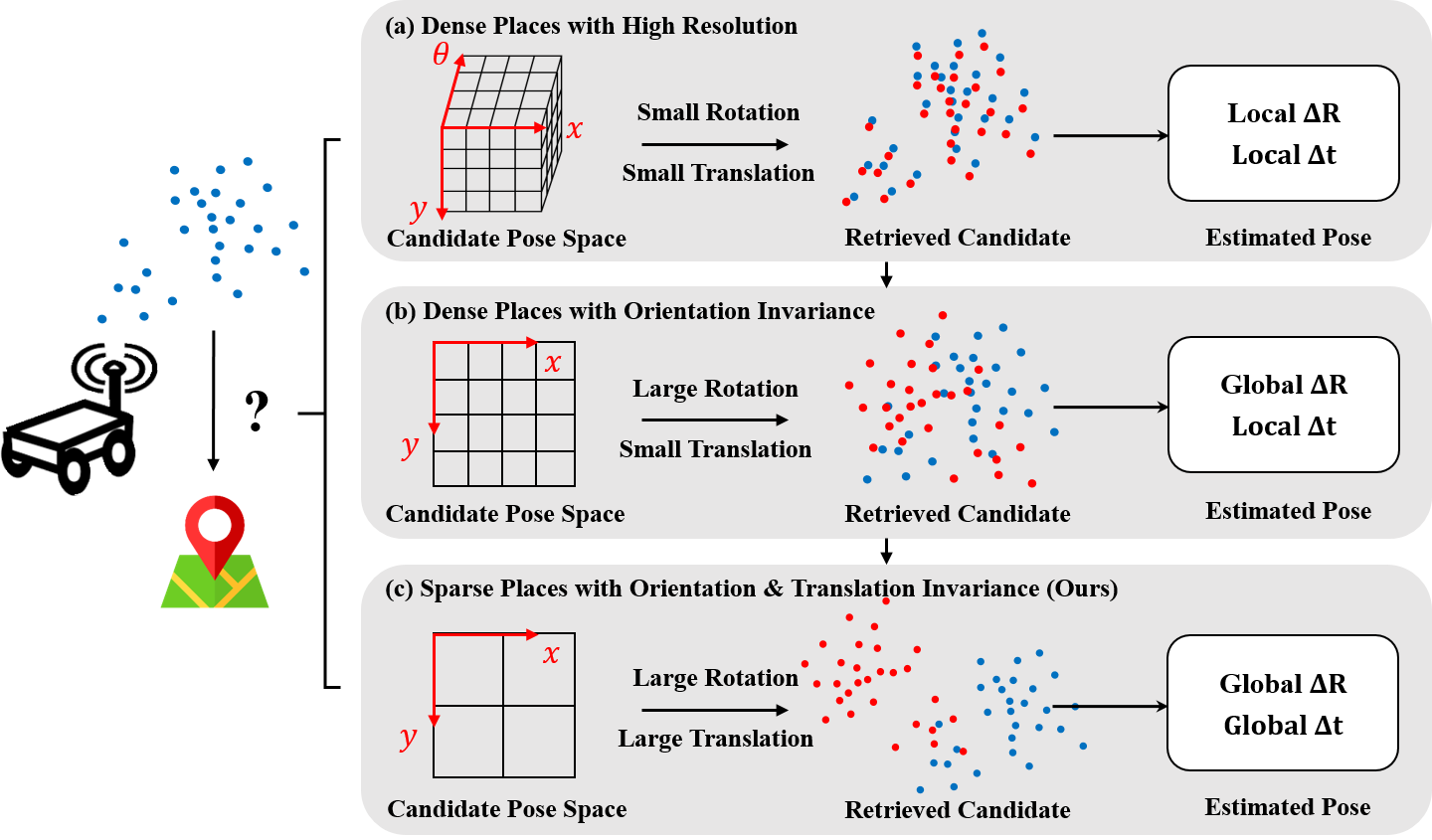}
\caption{Dense places and sparse places. We classify the global localization methods to two stages according to the map resolution. Our method aims at sparse places with orientation and translation invariance, as well as global convergent orientation and translation solvers.}
\label{fig:Dense vs. Sparse}
\vspace{-0.7cm}
\end{figure}

Imagining a spectrum aiming at this problem, one end represents \textit{dense places} with high resolution, which means only a small subset of poses is regarded as a place. It simplifies the pose estimation as local convergence can be sufficient in a small cell. However, this design calls for high computation and storage given the dense discretization in all dimensions as shown in block (a) of Fig. \ref{fig:Dense vs. Sparse}. On the contrary, the other end represents \textit{sparse places}, which infers a large area of the map, even the whole map, is regarded as one place. As a result, pose estimation becomes extremely difficult because it requires global convergence on the entire candidate pose space. Accordingly, the advantage of using sparse places is the lightweight computation and storage. In terms of global localization, we argue that only evaluation of place recognition may not fully reflect the performance of global localization. A more appropriate goal is to \textit{keep the pose estimation accuracy with a lower place density}.

Based on the analysis above, some early existing works follow the idea of dense places \cite{rohling2015fast,arandjelovic2016netvlad,yin2018locnet,pan2021coral}. These methods do not show very high computation. The reason is that robots in these scenarios usually travel along certain fixed trajectories, such as in-lane autonomous driving, which actually leads to reduced intrinsic dimensions of the candidate pose space. Nevertheless, in the case of general free space problems (the robot can move almost everywhere with arbitrary orientation, e.g., field robots in the wild), place recognition can be inefficient due to the irreducible candidate pose space. Therefore, recent studies like Scan Context \cite{kim2018scan} focuses on orientation invariance, which theoretically eliminates the dimension of orientation, leaving only the translation dimensions to be considered. These methods exponentially reduce the place density as shown in block (b) of Fig. \ref{fig:Dense vs. Sparse}. As a complement, global convergent orientation estimation is indispensable \cite{kim2018scan,chen2021overlapnet,xu2021disco,ding2022translation}. In this context, a further step to reduce the place density is to sparsify the translation space, which raises two critical challenges: place recognition robust to large relative orientation and translation, as well as efficient global convergence in both orientation and translation estimation.

Towards this goal, we propose a global localization solution as shown in block (c) of Fig. \ref{fig:Dense vs. Sparse}. We represent LiDAR scans in Radon sinogram (RING), which is very compact and allows for place recognition, orientation and translation estimation without additional feature engineering. The invariance of orientation and translation is theoretically derived for certifiable robustness and global convergence even under the large pose difference between current query scan and map scan, thus reducing the candidate pose space to sparse translation space. To the best of our knowledge, RING is the first learning-free unified representation to tackle all sub-tasks of global localization simultaneously. Experimental results validate RING in both place recognition and pose estimation, and show a superior global localization performance in more sparse place resolution. In summary, the contributions of this paper consist of:
\begin{itemize}
  \item
  A novel framework towards sparse place-based global localization, which utilizes a unified representation, RING, for all sub-tasks.
  \item
  Theoretical derivation on orientation and translation invariance to achieve certifiable robustness and global convergence.
  \item
  Evaluation of the proposed method on three large-scale multi-session datasets. The experimental results show the feasibility and superior performance even under lower place density. 
\end{itemize}

\section{Related Works}
For the place recognition task, numerous global point cloud descriptors have been proposed to represent the whole view of point cloud to achieve place description. These global descriptors can be categorized into handcrafted descriptors and deep learning descriptors. Some of them can solve pose estimation problem at the same time.

\subsection{Handcrafted Descriptors}
Early methods aggregate local features into a histogram and achieve the sparse encoding of a place. The common local features in computer vision community such as PFH \cite{rusu2008aligning}, SHOT \cite{salti2014shot} and spin images \cite{johnson1999using} are explored. Other than those matured local features in the community, Magnusson et al. \cite{magnusson2009appearance} utilized the surface orientation and smoothness to generate location histograms. Wohlkinger et al. \cite{wohlkinger2011ensemble} presented ESF which is a histogram of shape functions and Muhammad et al. \cite{muhammad2011loop} formed a histogram using normal vectors. Rohling et al. \cite{rohling2015fast} proposed a histogram to describe the range distribution of the entire point cloud. 

Other kinds of place recognition approaches rely on projections to reduce dimensionality. He et al. \cite{he2016m2dp} presented M2DP, which projects a LiDAR scan to multiple 2D planes and extracts the density signature of points in each plane as a global descriptor. To further achieve rotation-invariance of the global descriptor, Kim et al. \cite{kim2018scan} proposed a rotation-invariant descriptor called scan context image. They adopted polar transformation on the point clouds and used the height information as the feature in a polar bin. Recently, Kim et al. \cite{kim2021scan} further enhanced the scan context image to achieve robustness to lateral/rotational changes.

\subsection{Deep Learning Descriptors}
The last decade witnessed increasingly rapid progress in feature extraction and classification, mainly backed up by advances in the area of deep learning. Employing a neural network to learn a compact descriptor of the point cloud allows for implicitly encoding and thus have more attractions than the handcrafted descriptors. Yin et al. \cite{yin2018locnet,yin20193d} presented LocNet which firstly uses a 2D CNN after a handcrafted range histogram to enhance the place representation. Uy et al. \cite{uy2018pointnetvlad} proposed PointNetVLAD which is a combination of two neural networks, a feature extractor, PointNet \cite{qi2017pointnet} and a feature aggregator, NetVLAD \cite{arandjelovic2016netvlad}. Furthermore, based on the handcrafted descriptor, scan context image, Kim et al. \cite{kim20191} introduced a CNN as a classifier for solving long-term place recognition problem. Cramariuc et al. \cite{cramariuc2018learning} developed an approach that uses CNN to extract segments descriptors. 

Intend to cover more in global localization, Schaupp et al. \cite{schaupp2019oreos} proposed a system named OREOS for place recognition as well as orientation estimation. It uses a CNN as a feature extractor and a multilayer perceptron as an orientation estimator. Chen et al. \cite{chen2021overlapnet} further utilized correlation head after a CNN as an orientation estimator. Xu et al. \cite{xu2021disco} adopted Fourier transform and differentiable phase correlation to achieve explicit rotation invariance in global descriptors. Despite the powerful feature extraction and rotation estimation abilities of deep learning, there is a challenge in achieving translation invariance. In this paper, we introduce RING representation equipped with orientation and translation invariance, which addresses both place recognition and pose estimation problems at a lower place density.

\section{Preliminaries}
%In this section, we briefly introduce the properties of RING used in the proposed framework. 
% We also describe the generation of radon sinogram (RING) descriptor. The pipeline for place recognition and pose estimation with RING is presented as well.

\subsection{Radon Transform}

RING is the result of applying Radon transform to a 2D function $f(x,y)$. It is a integral transform of $f(x,y)$ along straight lines $L$, which projects a 2D function from the image space $(x,y)$ to the parameter space $(\theta,\tau)$. Therefore, RING is denoted by $\mathcal{R}_{f}(L)=\mathcal{R}_{f}(\theta, \tau)$.

Scanning lines $L$ are parameterized by $\theta$ and $\tau$ as:
\begin{equation}
x\cos\theta + y\sin\theta = \tau, (x,y)\in \mathbb{R}^2
\end{equation}
where $\theta$ is the incident angle between line $L$ with the $y$ axis and $\tau$ is the perpendicular distance from the origin to $L$. The range of $\theta$ is $[0, 2\pi)$ and the range of $\tau$ is $(-\infty, \infty)$. Then RING along a line is defined as
\begin{equation}
\begin{split}
& \mathcal{R}_{f}(\theta, \tau)=\int_{x\cos\theta + y\sin\theta = \tau} f(x,y) \mathrm{d}x \mathrm{d}y \\ &=\int_{-\infty}^{\infty}\int_{-\infty}^{\infty} f(x,y) \delta(\tau-x\cos\theta-y\sin\theta) \mathrm{d}x \mathrm{d}y
\end{split}
\end{equation}
where $\delta(\cdot)$ is the Dirac delta function.
\subsection{Basic Properties}
%\subsubsection{Periodicity}
%The Radon transform of $f(x,y)$ is periodic in the $\theta$ axis with the period of 2$\pi$.
%\begin{equation}
%{R}_{f}(\theta,\tau) = {R}_{f}(\theta+2n\pi,\tau), \forall n \in \mathbb{Z}
%\end{equation}

\textbf{Translation:} A translation of ${f(x,y)}$ by ${\Delta x}$ and ${\Delta y}$ in the image space results in a $\theta$ dependent shift $s$ on $\tau$ axis of RING:
\begin{equation}
{R}_{f}(\theta,\tau) \to {R}_{f}(\theta, \tau - \Delta x\cos\theta - \Delta y\sin\theta)	
\label{rttr}
\end{equation}
Note that translation in image space actually causes a non-uniform change in the RING. 

\textbf{Orientation:} An orientation of ${f(x,y)}$ by an angle $\alpha$ leads to a circular shift on $\theta$ axis of RING:
\begin{equation}
{R}_{f}(\theta,\tau) \to {R}_{f}(\theta + \alpha, \tau)				
\label{rtor}
\end{equation}
Compared to the image space translation, orientation change leads to uniform shift.

From (\ref{rttr}) and (\ref{rtor}), we can find that the translation and orientation of ${f(x,y)}$ is decoupled in the two variables $\tau$ and $\theta$ of RING. These properties help develop the invariance, which is beneficial for place recognition, as well as orientation and translation estimation.

\begin{figure}[t]
\centering
\includegraphics[width=8.5cm]{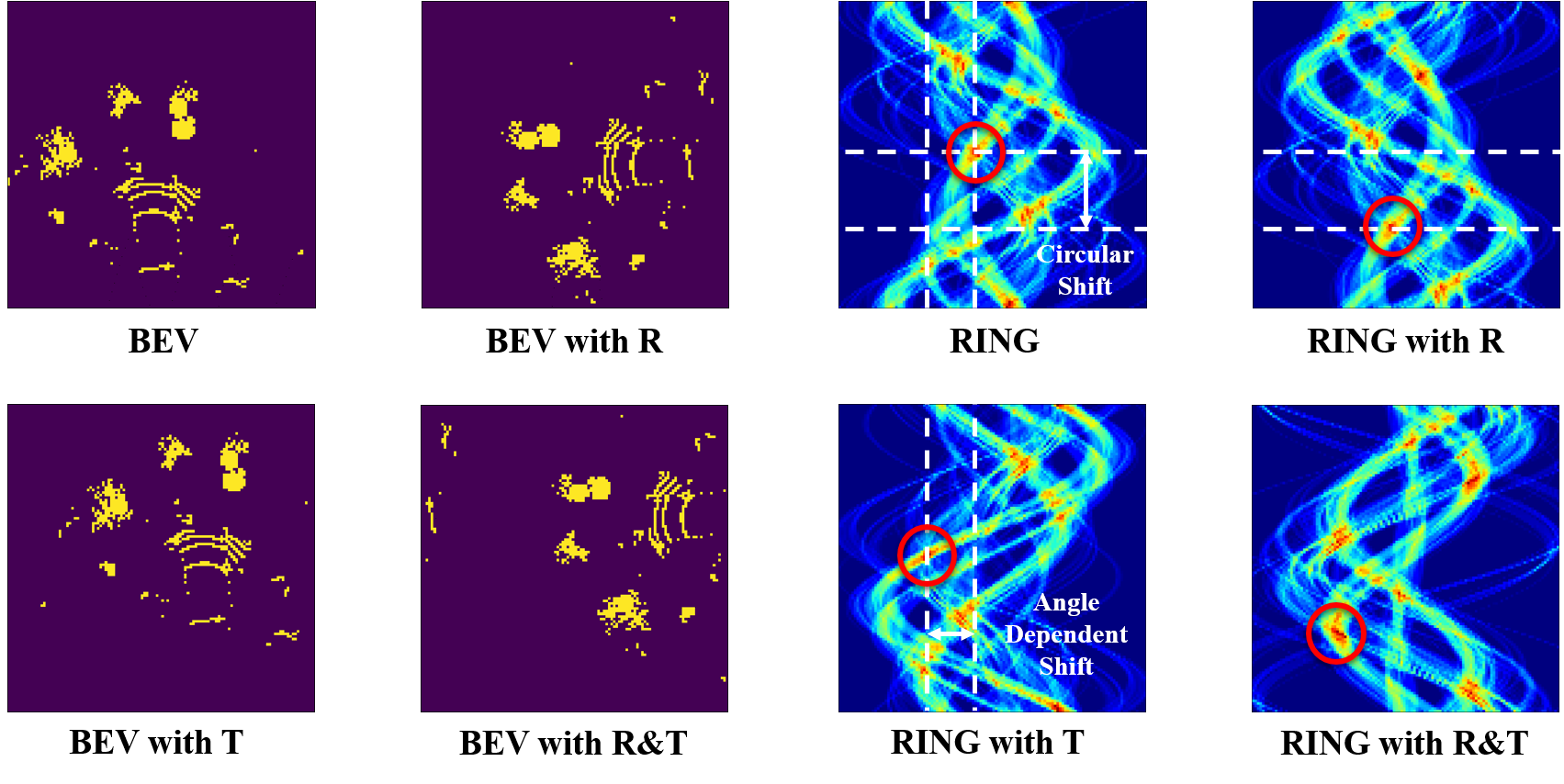}
\caption{Graphical illustration of orientation and translation properties of Radon Transform used in our method (R: Rotation, T: Translation). A rotation of point cloud corresponds to a circular column shift of RING. A translation of point cloud corresponds to an angle-dependent shift of RING.}
\label{fig:RING_Properties}
\vspace{-0.7cm}
\end{figure}

\section{Methodology}

\begin{figure*}
\centering
\includegraphics[width=18.2cm]{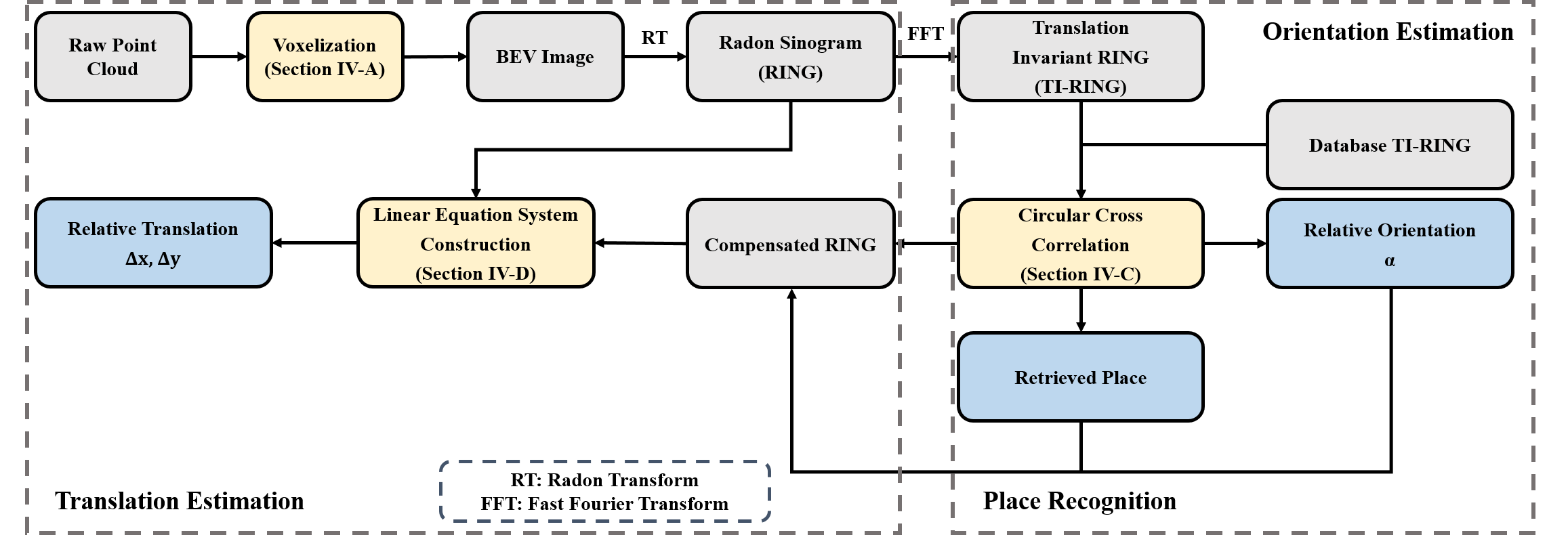}
\caption{Overview of the proposed framework about RING.}
\label{fig:overview}
\vspace{-0.7cm}
\end{figure*}

Given a query scan $P_Q$ and a database of map scans $\{P_D, T_D\}$ where $T_D$ is the pose of $P_D$, the problem of global localization is to calculate the pose $T_Q$ of $P_Q$. To address the problem, the proposed framework consists of three components. The LiDAR scan is firstly represented by RING. Then a translation invariant descriptor and an orientation invariant metric are built for place recognition, which retrieves a map scan $P_D$ that is close to where $P_Q$ is collected. Based on the retrieved map scan $P_D$, the relative orientation $\alpha$ and translation $[\Delta x, \Delta y]^T$ between $P_Q$ and $P_D$ is estimated, arriving at the final resultant global pose by compositing to $T_D$. Therefore, the goal is to keep the accuracy of estimated $\hat{\alpha}$ and $[\Delta \hat{x}, \Delta \hat{y}]^T$ by expanding the concept of `close'. The overview of our proposed framework is displayed in Fig. \ref{fig:overview}.

\subsection{RING based Scan Representation}

We follow the idea of Scan Context \cite{kim2018scan} to project the raw point cloud of a single scan into a bird-eye view representation (BEV). Instead of using the maximum height of point cloud within a BEV grid, we remove the ground from the point cloud and only use the binary occupancy to represent whether there is a point cloud within the grid. Such representation is lightweight, and not affected by the different absolute heights of sensor and environment.

\textbf{Volume to RING:} We begin with voxelizing the ground removed point cloud, where the height axis of the volume is aligned with the vertical axis of the LiDAR sensor. Then we sum the voxels along the height dimension to reduce the volume to the occupancy BEV, which is regarded as the 2D function $f(x,y)$. Then we apply radon transform to $f(x,y)$, yielding RING based scan representation ${R}_{f}(\theta,\tau)$. In the sequel, RING of query scan is denoted as ${R}_{Q}(\theta,\tau)$, RING of map scan is denoted as ${R}_{D}(\theta,\tau)$.

% The input data type of our algorithm is not limited to point clouds but also accepts 3D voxels and BEV images.

\subsection{Place Recognition}

Given a query scan ${P}_{Q}$ and a map scan ${P}_{D}$ sharing the same place, we have the relationship between ${R}_{Q}(\theta,\tau)$ and ${R}_{D}(\theta,\tau)$ according to (\ref{rttr}) and (\ref{rtor}):
\begin{equation}
{R}_{D}(\theta, \tau) = {R}_{Q}(\theta + \alpha, \tau - \Delta x\cos(\theta + \alpha) - \Delta y\sin(\theta + \alpha))
\end{equation}
In practice, we cannot know the relative orientation and translation between the query scan and arbitrary map scan. So we build a translation invariant descriptor for the scan, and an orientation invariant metric to measure whether the two scans are acquired from the same place. Due to the limited scan range and overlapping, in order to decrease the place density, the invariance is expected to reserve under arbitrary $\alpha$ and as large $\tau$ as possible.

% For successful place recognition, the place retrieval descriptors need to be invariant to arbitrary orientation so that the descriptors of query scan ${P}_{Q}$ and similar map scan ${P}_{D}$ can keep the same for any heading angle $\alpha$. Besides, the descriptors should be robust to translation change ($x$, $y$) to get rid of effect from motion disturbance.

\textbf{Translation invariant descriptor:}
In (\ref{rttr}), a translation in BEV is projected to an angle-dependent shift in the variable of $\tau$, which is shown in Fig. \ref{fig:RING_Properties}. To eliminate the effect of translation, we apply 1D discrete Fourier transform (DFT) on each row ($\tau$ axis) of RING and get the magnitude spectrum in the frequency domain. We define the resultant spectrum of RING as TI-RING. Denote query TI-RING and map TI-RING as ${M}_{Q}$ and ${M}_{D}$ respectively, the relationship between ${M}_{Q}$ and ${M}_{D}$ can be written as:
\begin{equation}
\begin{aligned}
&{M}_{D}(\theta, \omega)= |\mathcal{F}({R}_{D}(\theta, \tau))| \\
&= |\mathcal{F}({R}_{Q}(\theta + \alpha, \tau - \Delta x\cos(\theta + \alpha) - \Delta y\sin(\theta + \alpha)))| \\
&= |\mathcal{F}({R}_{Q}(\theta + \alpha, \tau))| = {M}_{Q}(\theta + \alpha, \omega)
\end{aligned}
\label{TIRING}
\end{equation}
where $\mathcal{F}(\cdot)$ is the DFT operator and $\omega$ is the discrete frequency. From (\ref{TIRING}), we notice that the relationship between ${M}_{Q}$ and ${M}_{D}$ is decoupled from the relative translation $\Delta x$ and $\Delta y$. Such exact translation invariance is only affected by the limit scan range.

\begin{figure}[t]
\centering
\includegraphics[width=7.5cm]{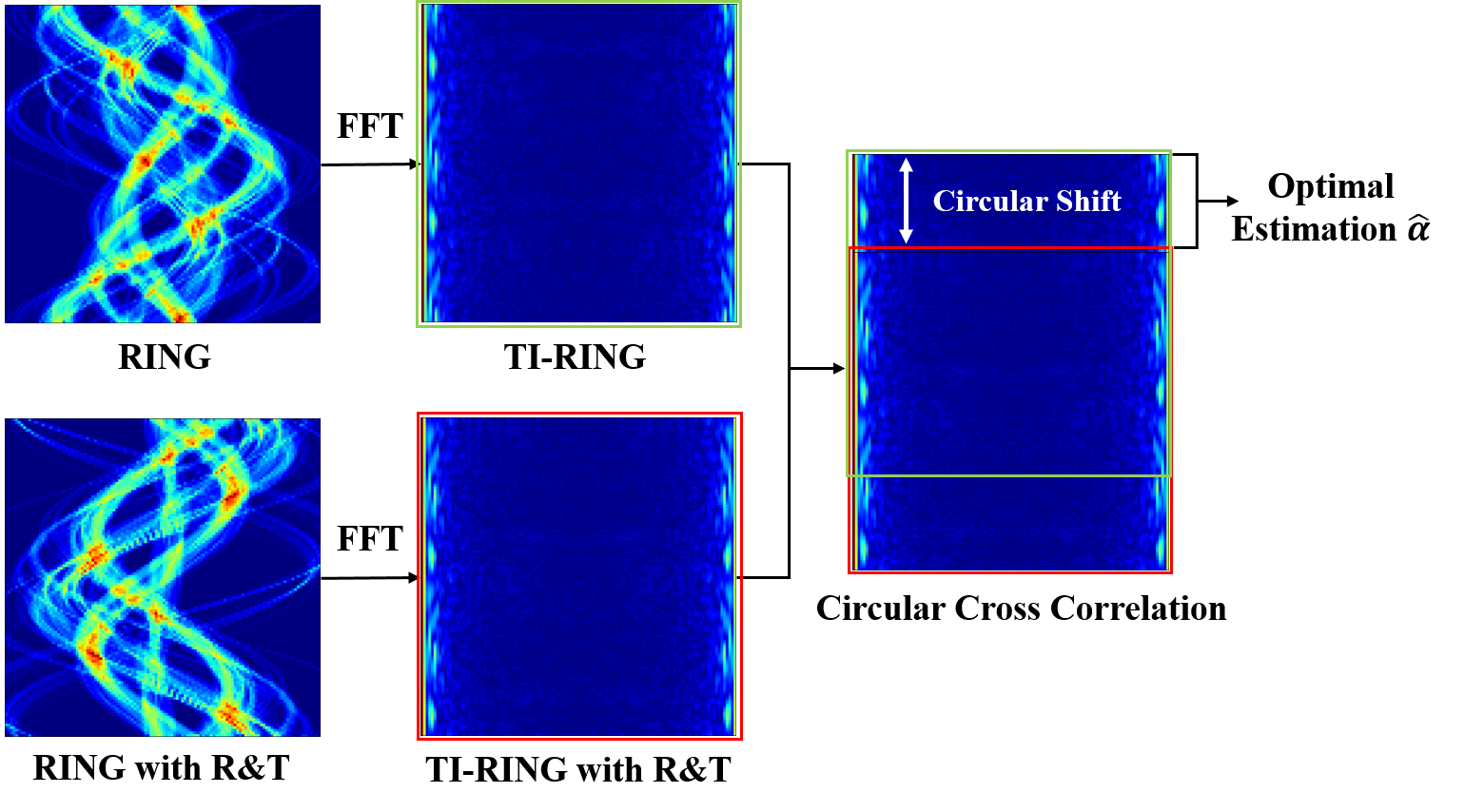}
\caption{Illustration of circular cross-correlation on TI-RING for orientation estimation (R: Rotation, T: Translation).}
\label{fig:Correlation}
\vspace{-0.55cm}
\end{figure}

\textbf{Orientation invariant similarity metric:}
In (\ref{rtor}), an orientation shift in BEV is converted to a circular row shift in $\theta$ axis and it is decoupled with translation, which is depicted in Fig. \ref{fig:RING_Properties} as well. In order to achieve orientation invariant metric for measuring similarity, we apply circular cross-correlation along the $\theta$ axis between ${M}_{Q}$ and ${M}_{D}$ as depicted in Fig. \ref{fig:Correlation}. The max correlation value of the circular cross-correlation results, denoted as $S({M}_{Q}, {M}_{D})$, is considered as the metric between the query scan ${P}_{Q}$ and map scan ${P}_{D}$:
\begin{equation}
S({M}_{Q}, {M}_{D}) = \max_\alpha \sum_{\theta_{i}} {M}_{Q}(\theta_{i}, \omega)\cdot{M}_{D}(\theta_{i}+\alpha, \omega)
\label{sim}
\end{equation}
where $\cdot$ is the inner product. To increase the efficiency, we utilize the GPU-based fast Fourier transform (FFT) to compute the circular cross-correlation. By comparing the query TI-RING with map TI-RING using (\ref{sim}), we can find a most similar map scan which should be acquired in the same place with query scan. 

\textbf{Presence of large translation:} Note that the TI-RING based similarity reserves the orientation invariance even larger relative translation presents, which is the main difference from previous methods that achieves orientation invariance with small relative translation \cite{kim2018scan,chen2021overlapnet,xu2021disco}. This is an important step towards sparse and efficient global localization. 

% Taking advantage of the shift property of Radon transform, we are able to fulfil place recognition, orientation estimation and translation estimation these three task together. 

\subsection{Orientation Estimation}
As shown in (\ref{TIRING}), the relative orientation estimation between ${P}_{Q}$ and ${P}_{D}$ is equivalent to estimate the $\theta$ axis shift between ${M}_{Q}$ and ${M}_{D}$, which can be completed by directly utilizing the TI-RING. Note that the optimal estimation $\hat{\alpha}$ of relative orientation is the one that achieves the maximum similarity metric (\ref{sim}):
\begin{equation}
\hat{\alpha} = \arg \max_{\alpha} \sum_{\theta_{i}} {M}_{Q}(\theta_{i}, \omega)\cdot{M}_{D}(\theta_{i}+\alpha, \omega)
\label{rotest}
\end{equation}

\textbf{Simultaneous solution:} Note that orientation estimation is a by-product when solving place recognition, thus is solved very efficiently. Besides, the orientation solver is global convergent since it estimates the relative orientation via exhaustive search with circular cross-correlation. Understandably, recognizing the correct place is coupled with estimating the correct orientation given that translation invariant descriptor is used.

\subsection{Translation Estimation}
After approximating the relative orientation $\hat{\alpha}$, we can use it to compensate the circular shift of map RING. The shifted map RING descriptor is denoted as ${R}^{'}_{D}(\theta, \tau)$. Theoretically, if we apply inverse Radon transform to ${R}^{'}_{D}(\theta, \tau)$, the resultant BEV should only have a translation from the query BEV. More specifically, the relationship between ${R}^{'}_{D}(\theta, \tau)$ and ${R}_{Q}(\theta, \tau)$ becomes:
\begin{equation}
{R}^{'}_{D}(\theta, \tau) = {R}_{Q}(\theta, \tau - \Delta x\cos(\theta + \alpha) - \Delta y\sin(\theta + \alpha))
\end{equation}

\begin{figure}[t]
\centering
\includegraphics[width=6.8cm]{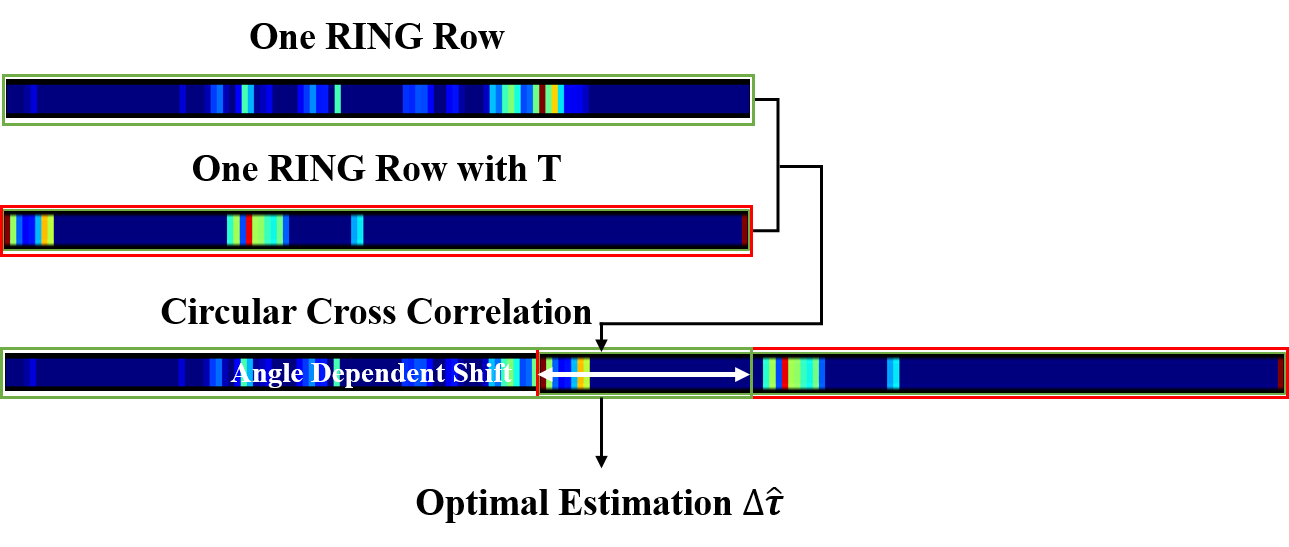}
\caption{Illustration of circular cross-correlation on RING for translation estimation (T: Translation).}
\label{fig:Correlation_Row}
\vspace{-0.8cm}
\end{figure}

\textbf{Linear equation system construction:} Given the correct approximation i.e. $\hat{\alpha} \approx\alpha$, for each row pair of ${R}^{'}_{D}(\theta, \tau)$ and ${R}_{Q}(\theta, \tau)$, we can get a linear equation according to (\ref{rttr}) with unknowns $\Delta x$ and $\Delta y$:
\begin{equation}
\Delta x\cos(\theta + \alpha) + \Delta y\sin(\theta + \alpha) = \Delta \hat{\tau}
\end{equation}
where $\Delta \hat{\tau}$ is the optimal estimated shift between the paired row using the circular cross-correlation implemented by FFT as shown in Fig. \ref{fig:Correlation_Row}. Assume the number of columns of a RING descriptor is $k$ ($k = 120$ in our experiment). Then we have $k$ linear equations with $\Delta x$ and $\Delta y$ as unknowns, which is an over-determined linear equation system:
\begin{equation}
\left[\begin{array}{cc}
\cos (\theta_{1} + \alpha) & \sin (\theta_{1} + \alpha)  \\
\cos (\theta_{2} + \alpha)  & \sin (\theta_{2} + \alpha)  \\
\vdots & \vdots \\
\cos (\theta_{k} + \alpha)  & \sin (\theta_{k} + \alpha) 
\end{array}\right]\left[\begin{array}{l}
\Delta x \\
\Delta y
\end{array}\right]=\left[\begin{array}{c}
\Delta \hat{\tau}_{1} \\
\Delta \hat{\tau}_{2} \\
\vdots \\
\Delta \hat{\tau}_{k}
\end{array}\right]
\label{les}
\end{equation}
Note that $\Delta \hat{\tau}$ varies for different $\theta_k$. The noise in both estimations of $\alpha$ and $\Delta \tau$ can affect the equation coefficients, which leads to error in translation estimation.

\textbf{Closed form:} The translation estimation problem is converted to solve a linear equation system (\ref{les}), which can be calculated in closed form, arriving at $[\Delta \hat {x}, \Delta \hat{y}]^T$. 

% The angle-dependent shift between ${M}_{Q}$ and ${M}_{D}$ caused by translation can be used to solve the relative translation.

\section{Experimental Results}

In the experiments, we verify the performance of place recognition and pose estimation using the proposed method:
\begin{itemize}
\item validate the orientation and translation invariance of TI-RING by testing the performance of place recognition at varied place density.
\item validate the accuracy of pose estimation when the relative orientation and translation between the map and query scan are large.
\item compare with other methods in terms of global localization success rate with respect to the place density.
\end{itemize}

\subsection{Dataset and Experimental Settings}

We utilize three large-scale datasets for evaluation: NCLT Dataset \cite{carlevaris2016university}, MulRan Dataset \cite{kim2020mulran} and Oxford Radar RobotCar Dataset \cite{barnes2020oxford}. These datasets contain multiple sessions collected in different positions and orientations across various environmental conditions. 
% orientation (revisit at reverse direction), lane (at two different lateral lanes)

\textbf{NCLT Dataset:} NCLT Dataset is a large-scale and long-term dataset collected by a Segway robot at the University of Michigan North Campus. % The dataset is collected to facilitate research focusing on long-term autonomous operation in changing environments. 
It consists of 27 sessions across four seasons, spaced approximately every two weeks for 15 months. We use the trajectory collected on "2012-02-04" as map data and other trajectories as query data for evaluation.

\textbf{MulRan Dataset:} MulRan Dataset comprises multimodal range data: Radar and LiDAR for robust structural place recognition algorithms. It is collected in multiple cities in different environments. In our experiments, we choose sessions from DCC and KAIST for place recognition evaluation. 

\textbf{Oxford Radar RobotCar Dataset:} Oxford Radar RobotCar Dataset provides optimised ground truth collected by Radar and two Velodyne HDL-32E LiDARs. We concatenate point clouds collected by the left and right LiDARs mounted on the vehicle to one single point cloud for evaluation.

On all datasets, the ROI of the LiDAR scan is cropped to $140m\times140m$ centering at the sensor frame, and is represented by a $120\times 120$ BEV with a resolution of $1.17m/pixel$ for RING extraction.

% \begin{table}[htbp]
% \renewcommand\arraystretch{1.2}
% \centering
% \caption{Experimental Settings.}
% \label{table1}
% \begin{tabular}{c|c|c}
% \hline
% \hline
% Parameter & RING & TI-RING \\ \hline
% ROI & [-70m, 70m] & [0\degree, 360\degree] \\ \hline
% Resolution & 120$\times$120 & (1.17m, 3\degree) \\
% \hline
% \hline
% \end{tabular}
% \end{table}

\subsection{Comparative Methods}
To verify the translation invariance of TI-RING, we perform place recognition merely based on RING (i.e., we employ circular cross-correlation directly to RING rather than TI-RING) as a variant of our TI-RING method, denoted as RING method. We mainly compare RING and TI-RING methods with four handcrafted methods: M2DP \cite{he2016m2dp}, Fast Histogram \cite{rohling2015fast}, Scan Context \cite{kim2018scan}, and Scan Context++ \cite{kim2021scan}. All methods are implemented using Python. The input of M2DP and Fast Histogram is the raw point cloud. All parameters used in our experiments are the same as those in the original papers. The M2DP descriptor is a 192-length descriptor. The bucket count of Fast Histogram is set as 100. For a fair comparison, the resolution of SC and SC++ is the same as RING, which is $120 \times 120$. The number of candidates of Scan Context is 50, keeping the same as the original paper.

\subsection{Place Recognition Evaluation}
% We perform place recognition on three large-scale datasets mentioned above. 
We first evaluate our approach and other approaches on place recognition performance with respect to different revisit criteria and place density. By changing the revisit threshold and place density, the translation and orientation invariance of the descriptor can be reflected.

\begin{figure}[t]
\centering
\includegraphics[width=7.45cm]{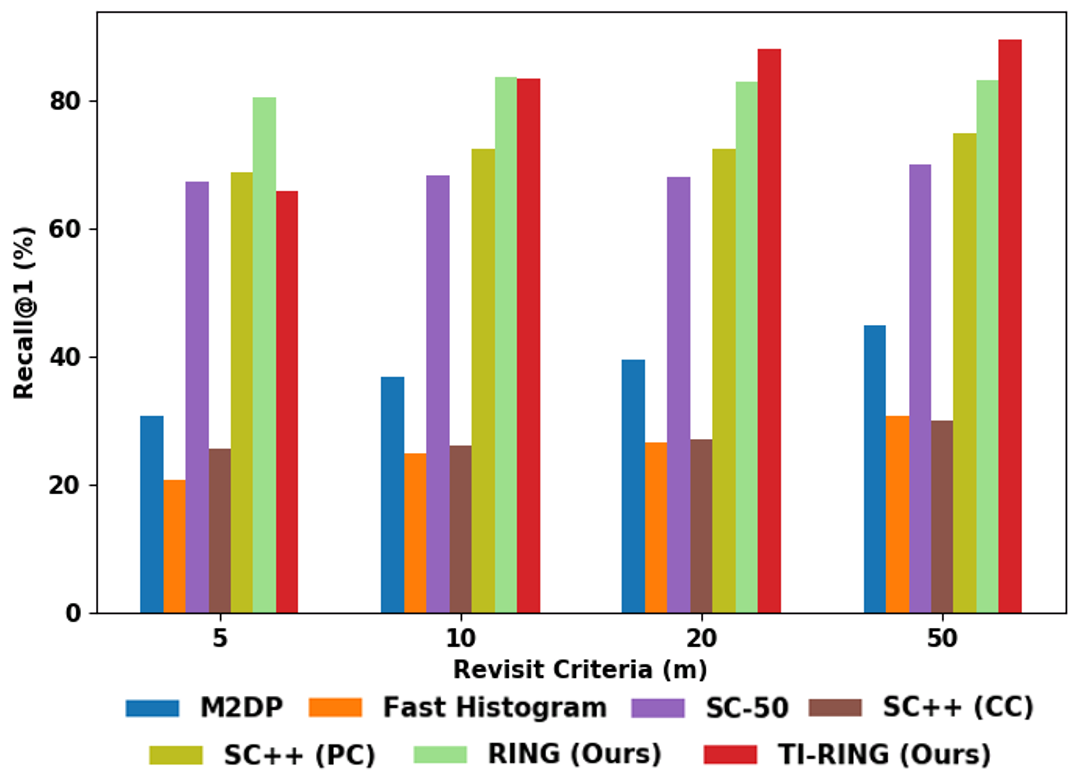}
\caption{Recall@1 vs. Revisit Criteria on NCLT dataset.}
\label{fig:Recall1}
\vspace{-0.8cm}
\end{figure}

\textbf{Revisit criteria:}
Revisit criteria is the threshold determining whether the query pose and the map pose belong to the same place. To verify the invariance of our method, we first perform place recognition on two trajectories of NCLT dataset with respect to varied revisit criteria. We take "2012-02-04" sequence as map data and "2012-03-17" as query data. The map scans and query scans are sampled at fixed and equidistant intervals (map: $10m$, query: $5m$). The revisit criteria range from $5m$ to $50m$. Recall@1 is used as the evaluation metric shown in Fig. \ref{fig:Recall1}.

% \begin{figure}[t]
% \centering
% \includegraphics[width=8cm]{Recall1.eps}
% \caption{Recall@1 vs. Revisit Criteria on NCLT dataset.}
% \label{fig:Recall1}
% \vspace{-0.5cm}
% \end{figure}

We notice that RING has competitive performance on all thresholds. The trend with respect to the revisit criteria is slightly smooth, which is similar to SC class methods. These results infer that the incorrectly recognized places are far away from the query pose. The trend of TI-RING is different, which lies a little lower than SC and RING under small threshold, but grows obviously with the increase of the threshold, and finally becomes the top. The underlying reason is the strong translation invariance of TI-RING, which is able to recognize places in a larger neighborhood. The other methods show lower performances, which may be explained by the discrimination capacity of the descriptors. Based on this result, we consider that our method is capable of handling place recognition with a lower place density.

\begin{figure}[t]
\centering
\includegraphics[width=7.8cm]{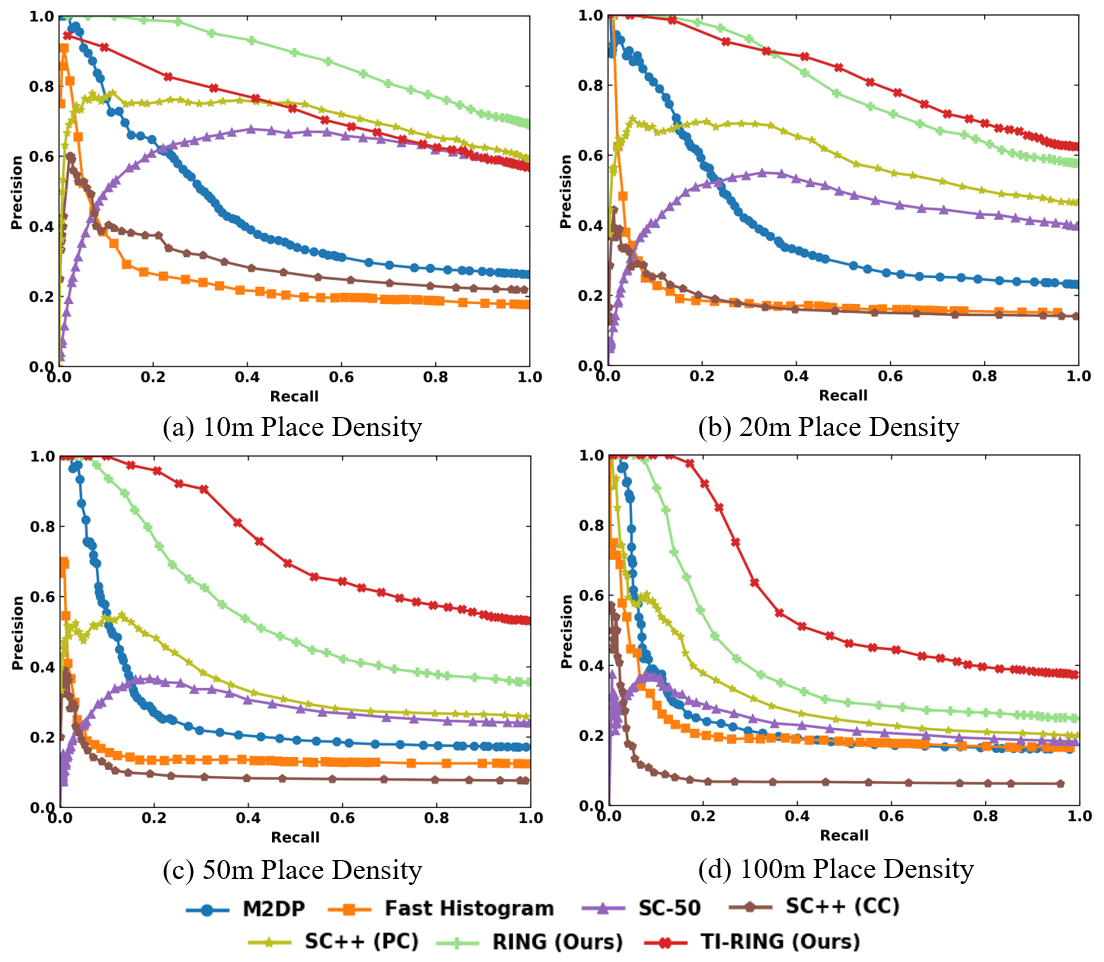}
\caption{Precision-recall curves at different map density on "2012-03-17" to "2012-02-04" for NCLT dataset.}
\label{fig:pr_map_sparsity}
\vspace{-0.5cm}
\end{figure}

\textbf{Place density:}
We furthermore perform place recognition on NCLT dataset by changing the pose density i.e. the sampling interval along the map trajectory. The density is set to $10m$, $20m$, $50m$ and $100m$ respectively. All query data is sampled at a fixed $5m$ interval. The revisit threshold is taken as one half of the density, which is $5m$, $10m$, $25m$ and $50m$. The results in Fig. \ref{fig:pr_map_sparsity} show that the performance of RING and TI-RING are a little higher than other methods at a density of $10m$. With the increase of density, our methods still maintain the top performance. While the performance of compared methods steeply drops, TI-RING, which is equipped with the translation invariance, only has a small drop from $0.6$ to $0.4$ at $100\%$ recall. This result further validates that our method has strong robustness to larger relative translation between query and map poses. 

\textbf{Cross-dataset consistency:}
To validate the consistency of the performance, we further compare the methods on all three datasets with a place density of $50m$. For NCLT dataset, we select "2012-02-04" sequence as map data, "2012-05-26" and "2012-08-20" sequences as query data to evaluate place recognition performance. For MulRan dataset, we select "DCC01"/"KAIST01" trajectory as map data, "DCC02"/"KAIST02" trajectory as query data for evaluation. These sequences are relatively simple as low orientation and translation variance presents. For Oxford dataset, we choose "2019-01-15-13-06-37" to "2019-01-11-13-24-51" and "2019-01-16-11-53-11" to "2019-01-11-13-24-51" two pairs. In this part, we choose precision-recall curve and F1 score-recall curve as the evaluation metric. Results with a place density of $50m$ are depicted in Fig. \ref{fig:pr_nclt} to Fig. \ref{fig:pr_oxford}. 
Results with a place density of $20m$ are shown in Appendix \cite{lusha2022full}.
Compared with other methods, TI-RING shows an obviously better performance, followed by RING, which is consistent on all datasets. In Table \ref{table2}, we summarize the place recognition performance on Oxford dataset with two place density settings, $20m$ and $50m$. Accuracy and F1 score are measured at $100\%$ recall for comparison. The evaluated values of TI-RING are almost twice as the other methods for $50m$ place density. Moreover, the performance of our approach at $50m$ place density is almost $30\%$ better than the best comparative methods at a smaller $20m$ place density, showing great improvement in sparse places.

% \begin{figure}[t]
% 	\centering
% 	\subfigure[10m Place density]{
% 		\includegraphics[width=4.12cm]{PR_5m.eps}}
% 	\subfigure[20m Place density]{
% 		\includegraphics[width=4.12cm]{PR_10m.eps}}
% 	\subfigure[50m Place density]{
% 		\includegraphics[width=4.12cm]{PR_25m.eps}}
%     \subfigure[100m Place density]{
% 		\includegraphics[width=4.12cm]{PR_50m.eps}}
% 	\caption{Precision-recall curves at different map density on two sessions of NCLT dataset: 2012-02-04 (map) to 2012-03-17 (query).}
% 	\label{fig:pr_map_sparsity}
%     \vspace{-0.5cm}
% \end{figure}

\begin{figure}[!t]
	\centering
	\includegraphics[width=7.8cm]{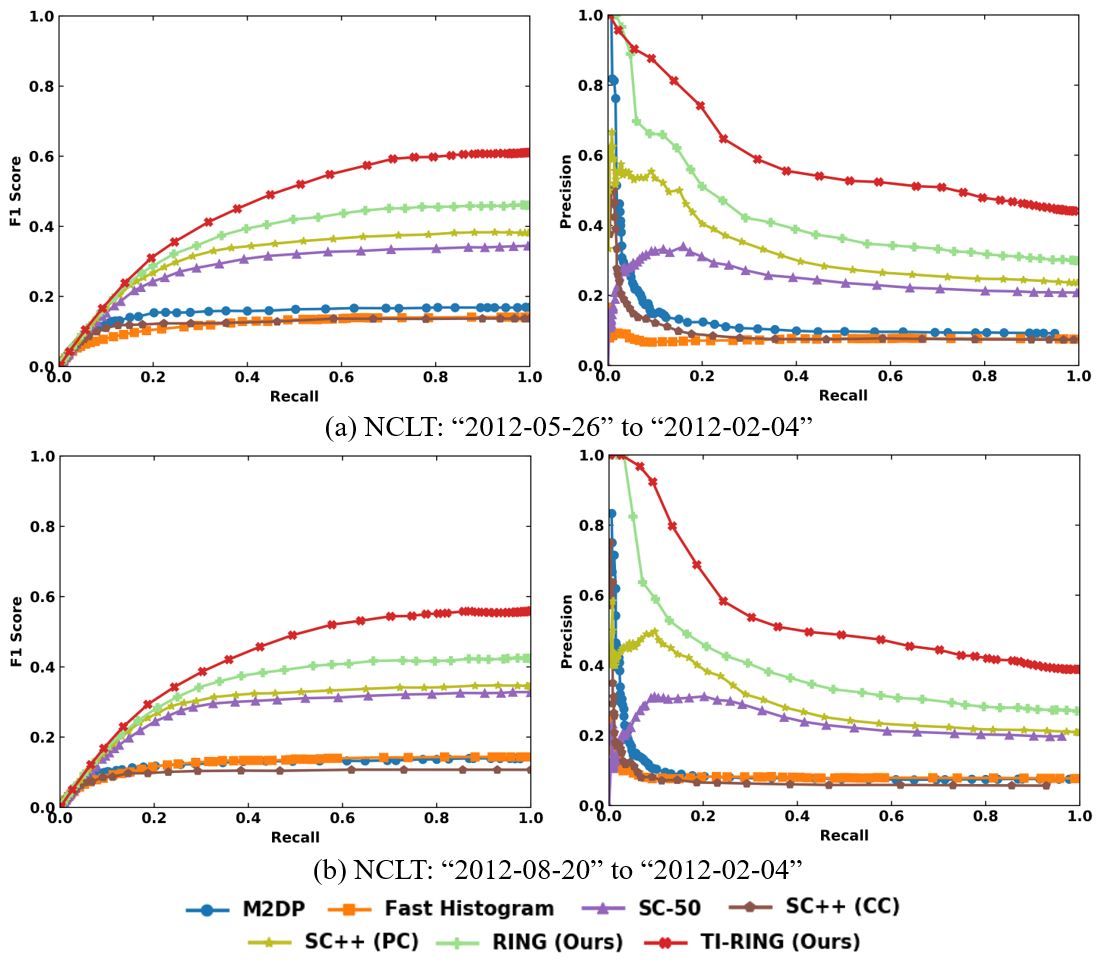}
	\caption{F1 score-recall (left) and precision-recall (right) curves on NCLT dataset (50m place density).}
	\label{fig:pr_nclt}
    \vspace{-0.8cm}
\end{figure}

\begin{figure}[t]
	\centering
    \includegraphics[width=7.8cm]{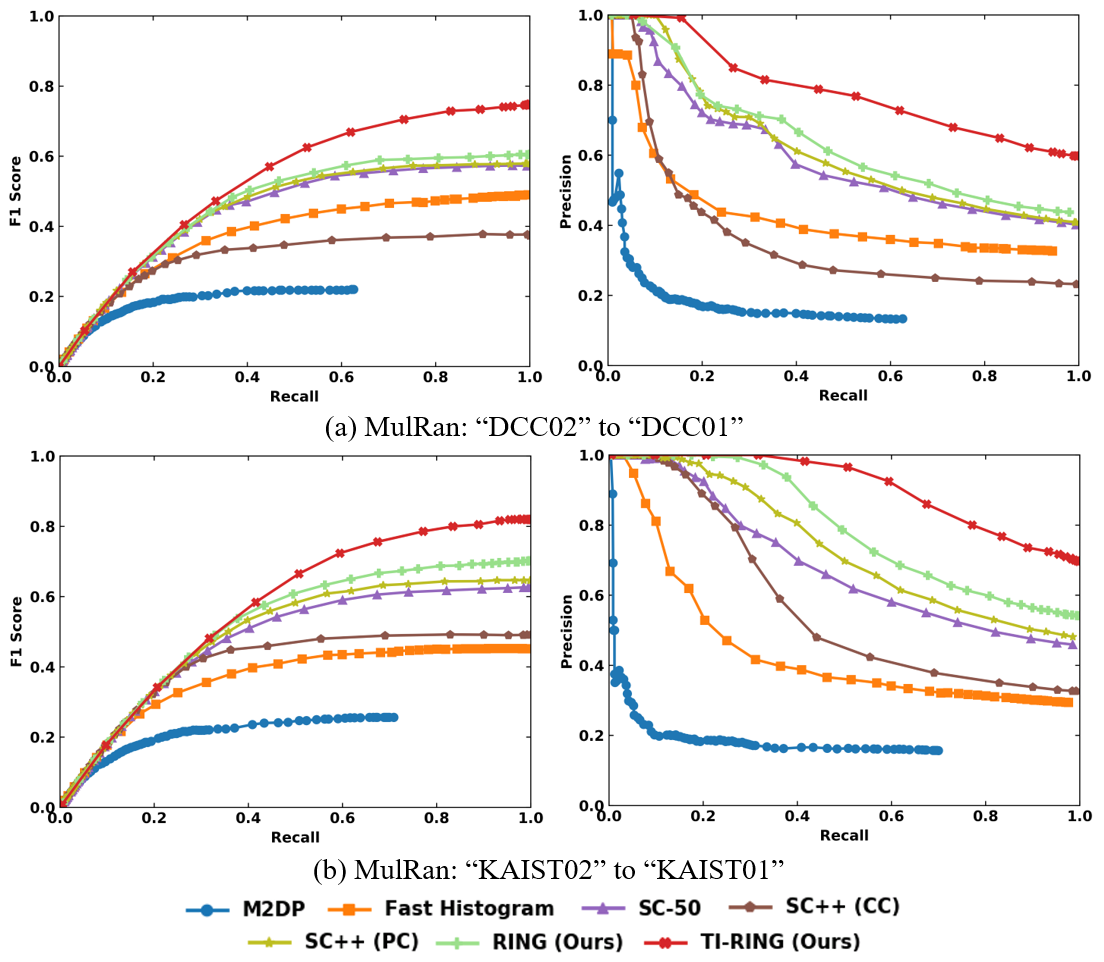}
	\caption{F1 score-recall (left) and precision-recall (right) curves on MulRan dataset (50m place density).}
	\label{fig:pr_mulran}
    \vspace{-0.5cm}
\end{figure}

\begin{figure}[!t]
	\centering
    \includegraphics[width=7.8cm]{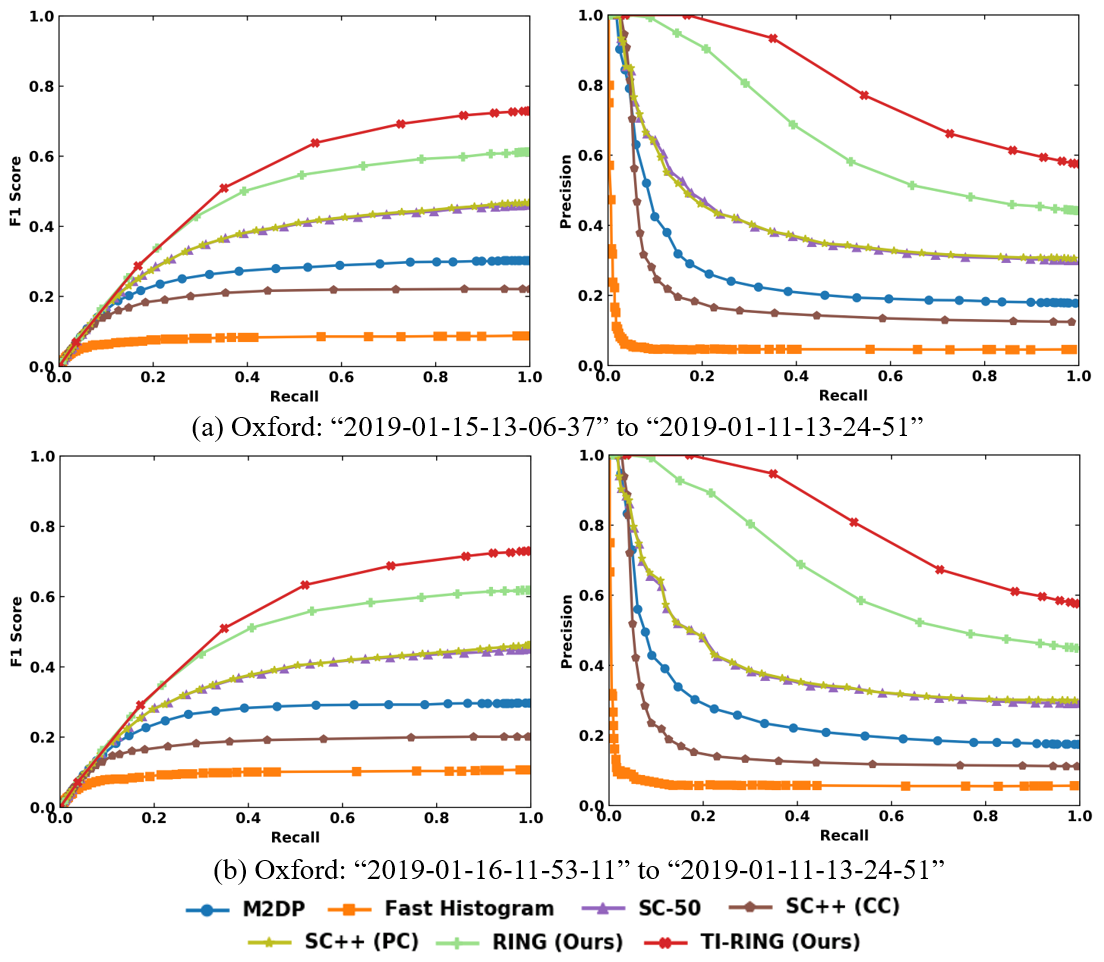}
	\caption{F1 score-recall (left) and precision-recall (right) curves on Oxford dataset (50m place density).}
	\label{fig:pr_oxford}
    \vspace{-0.8cm}
\end{figure}

\begin{table}[htbp]
\renewcommand\arraystretch{1.2}
\centering
\caption{Place recognition performance on Oxford dataset}
\label{table2}
\begin{tabular}{clccc}
\hline
\textbf{Place Density} & \textbf{Approach} & \textbf{Accuracy} & \textbf{F1 score} & \textbf{AUC} \\ \hline
& \textbf{TI-RING (Ours)} & \textbf{0.7812} & \textbf{0.8772} & \textbf{0.9234} \\ 
& RING (Ours) & 0.7191 & 0.8366 & 0.8787 \\
& M2DP & 0.2900 & 0.4496 & 0.4638 \\
20m& Fast Histogram & 0.0628 & 0.1181 & 0.0741 \\ 
& SC-50 & 0.4220 & 0.5935 & 0.5366 \\ 
& SC++ (PC) & 0.4252 & 0.5967 & 0.5424\\ 
& SC++ (CC) & 0.2485 & 0.3981 & 0.3710 \\ \hline
& \textbf{TI-RING (Ours)} & \textbf{0.5741} & \textbf{0.7295} & \textbf{0.8090} \\ 
& RING (Ours) & 0.4421 & 0.6131 & 0.6622 \\
& M2DP & 0.1780 & 0.3022 & 0.2650 \\
50m& Fast Histogram & 0.0460 & 0.0879 & 0.5312 \\
& SC-50 & 0.2990 & 0.4604 & 0.4092 \\ 
& SC++ (PC) & 0.3049 & 0.4673 & 0.4096\\ 
& SC++ (CC) & 0.1243 & 0.2211 & 0.1973 \\  
\hline
\end{tabular}
\vspace{-0.15cm}
% \vspace{-0.7cm}
\end{table}

\subsection{Pose Estimation Evaluation}
After place recognition, the downstream task is the pose estimation to gain the current robot pose. To perform pose estimation analysis, we choose NCLT dataset as the evaluation dataset ("2012-02-04" sequence acts as map data, other sequences act as query data). Among the compared methods, only SC and SC++ can estimate orientation. Therefore, we compare our method with SC and SC++. According to the pipeline proposed in Fig. \ref{fig:overview}, we estimate the relative orientation between two scans first. Then we compute the translation after compensating the orientation on RING. In this experiment, the pose estimation accuracy is only evaluated on scans of correctly recognized places.

% \begin{figure*}[htbp]
% 	\centering
%     \subfigure[Orientation Error]{
% 		\includegraphics[width=5.4cm]{yaw_error_boxplot.eps}}
% 	\subfigure[Lateral Translation Error]{
% 		\includegraphics[width=5.4cm]{lateral_error_boxplot.eps}}
%     \subfigure[Longitudinal Translation Error]{
% 		\includegraphics[width=5.4cm]{vertical_error_boxplot.eps}}
% 	\caption{Orientation and Translation Estimation Error on NCLT dataset.}
% 	\label{fig:Pose_Error}
% \end{figure*}
\begin{figure*}[htbp]
	\centering
	\includegraphics[width=14.5cm]{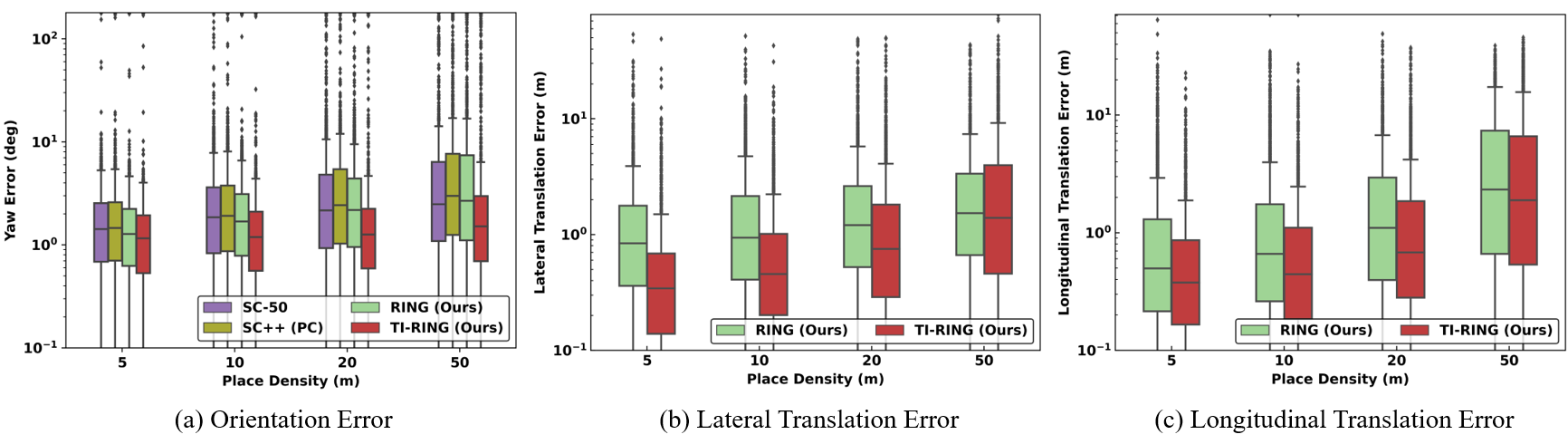}
	\caption{Orientation and translation estimation error on NCLT dataset.}
    \vspace{-0.8cm}
	\label{fig:Pose_Error}
\end{figure*}

\textbf{Orientation and translation:}
The overall pose estimation results with respect to varied place density are shown in Fig. \ref{fig:Pose_Error}. For the orientation estimation, TI-RING has the lowest orientation estimation error. With the increase of place density, the estimation errors of all methods become larger while the pose estimation errors of TI-RING increase much slower than other methods. These results can be explained that the effect of relative translation is theoretically eliminated for TI-RING. For the lateral and longitudinal translation estimation, TI-RING demonstrates accuracy in about sub-meter level with place density lower than $20m$. The inferior performance of RING is caused by the error of orientation estimation.

\textbf{Stratified global localization success:}
Following the proposed goal of keeping the accuracy with a lower place density, we evaluate the accuracy by the number of all query scans that are correctly localized on the map. The threshold of correctness is set as $3\degree$ and $3m$. For SC and SC++, the relative translation is assigned by $0$, which follows common practices that assigns the current pose by the pose of the recognized place. 
% Success rates on each of the four datasets are shown in Appendix \cite{lusha2022full}. 
By aggregating the results from all datasets, a final performance comparison is shown in Table \ref{paspd}. One can see that our method, TI-RING, rules place recognition, orientation estimation, and translation estimation at the same time. The margin over the comparative methods is obvious. Even with a place density of $50m$, TI-RING outperforms the others with a place density of $20m$, which means a superior accuracy is achieved with a smaller map.

\begin{table}[t]
\renewcommand\arraystretch{1.2}
\centering
\caption{Localization success stratified by place density}
\label{paspd}
\begin{threeparttable}
\begin{tabular}{clccc}
\hline
\makecell[c]{\textbf{Place Density}} & \textbf{Approach} & \makecell[c]{\textbf{TSR}} & \makecell[c]{\textbf{OSR}} & \makecell[c]{\textbf{LSR}} \\ \hline
 \multirow{4}{*}{20m} & SC-50 & 0.2418 & 0.2364 & 0.0729 \\
 & SC++ (PC) & 0.2382 & 0.2320 & 0.0667 \\
 & RING (Ours) & 0.3404 & 0.3360 & 0.2373 \\
 & \textbf{TI-RING (Ours)} & \textbf{0.5333} & \textbf{0.4453} & \textbf{0.3982} \\ \hline
  \multirow{4}{*}{50m}  & SC-50 & 0.1493 & 0.0827 & 0.0409 \\ 
 & SC++ (PC) & 0.1360 & 0.0800 & 0.0373 \\
 & RING (Ours) & 0.1822 & 0.1413 & 0.1049 \\
 & \textbf{TI-RING (Ours)} & \textbf{0.3956} & \textbf{0.2489} & \textbf{0.2258} \\
\hline
\end{tabular}
\begin{tablenotes}
    \footnotesize
    \item[*] TSR: Translation Success Rate, OSR: Orientation Success Rate, LSR: Localization Success Rate.
\end{tablenotes}
\end{threeparttable}
\vspace{-0.7cm}
\end{table}

\section{Conclusion}

In this paper, we propose a unified descriptor named RING as the representation of a place. The RING descriptor can achieve certifiable orientation and translation invariance, resulting in further reduction the resolution of candidate map pose space. In the experiments, our method shows notable performance on place recognition, especially applied at lower place density. Based on the retrieved places, pose estimation can be further achieved at a relatively small error.


\printbibliography

@article{lynen2020large,
  title={Large-scale, real-time visual--inertial localization revisited},
  author={Lynen, Simon and Zeisl, Bernhard and Aiger, Dror and Bosse, Michael and Hesch, Joel and Pollefeys, Marc and Siegwart, Roland and Sattler, Torsten},
  journal={The International Journal of Robotics Research},
  volume={39},
  number={9},
  year={2020},
  publisher={SAGE Publications Sage UK: London, England}
}

@article{lowry2015visual,
  title={Visual place recognition: A survey},
  author={Lowry, Stephanie and S{\"u}nderhauf, Niko and Newman, Paul and Leonard, John J and Cox, David and Corke, Peter and Milford, Michael J},
  journal={IEEE Transactions on Robotics},
  volume={32},
  number={1},
  pages={1--19},
  year={2015},
  publisher={IEEE}
}

@inproceedings{rohling2015fast,
  title={A fast histogram-based similarity measure for detecting loop closures in 3-d lidar data},
  author={R{\"o}hling, Timo and Mack, Jennifer and Schulz, Dirk},
  booktitle={2015 IEEE/RSJ International Conference on Intelligent Robots and Systems (IROS)},
  pages={736--741},
  year={2015},
  organization={IEEE}
}

@inproceedings{magnusson2009appearance,
  title={Appearance-based loop detection from 3D laser data using the normal distributions transform},
  author={Magnusson, Martin and Andreasson, Henrik and Nuchter, Andreas and Lilienthal, Achim J},
  booktitle={2009 IEEE International Conference on Robotics and Automation},
  year={2009},
  organization={IEEE}
}

@inproceedings{he2016m2dp,
  title={M2DP: A novel 3D point cloud descriptor and its application in loop closure detection},
  author={He, Li and Wang, Xiaolong and Zhang, Hong},
  booktitle={2016 IEEE/RSJ International Conference on Intelligent Robots and Systems (IROS)},
  pages={231--237},
  year={2016},
  organization={IEEE}
}

@inproceedings{kim2018scan,
  title={Scan context: Egocentric spatial descriptor for place recognition within 3d point cloud map},
  author={Kim, Giseop and Kim, Ayoung},
  booktitle={2018 IEEE/RSJ International Conference on Intelligent Robots and Systems (IROS)},
  pages={4802--4809},
  year={2018},
  organization={IEEE}
}

@article{yin20193d,
  title={3d lidar-based global localization using siamese neural network},
  author={Yin, Huan and Wang, Yue and Ding, Xiaqing and Tang, Li and Huang, Shoudong and Xiong, Rong},
  journal={IEEE Transactions on Intelligent Transportation Systems},
  volume={21},
  number={4},
  pages={1380--1392},
  year={2019},
  publisher={IEEE}
}

@inproceedings{pan2021coral,
  title={Coral: Colored structural representation for bi-modal place recognition},
  author={Pan, Yiyuan and Xu, Xuecheng and Li, Weijie and Cui, Yunxiang and Wang, Yue and Xiong, Rong},
  booktitle={2021 IEEE/RSJ International Conference on Intelligent Robots and Systems (IROS)},
  pages={2084--2091},
  year={2021},
  organization={IEEE}
}

@article{ding2022translation,
  title={Translation Invariant Global Estimation of Heading Angle Using Sinogram of LiDAR Point Cloud},
  author={Ding, Xiaqing and Xu, Xuecheng and Lu, Sha and Jiao, Yanmei and Tan, Mengwen and Xiong, Rong and Deng, Huanjun and Li, Mingyang and Wang, Yue},
  journal={arXiv preprint arXiv:2203.00924},
  year={2022}
}

@inproceedings{wohlkinger2011ensemble,
  title={Ensemble of shape functions for 3d object classification},
  author={Wohlkinger, Walter and Vincze, Markus},
  booktitle={2011 IEEE international conference on robotics and biomimetics},
  pages={2987--2992},
  year={2011},
  organization={IEEE}
}

@inproceedings{muhammad2011loop,
  title={Loop closure detection using small-sized signatures from 3D LIDAR data},
  author={Muhammad, Naveed and Lacroix, Simon},
  booktitle={2011 IEEE International Symposium on Safety, Security, and Rescue Robotics},
  year={2011},
  organization={IEEE}
}

@inproceedings{rusu2008aligning,
  title={Aligning point cloud views using persistent feature histograms},
  author={Rusu, Radu Bogdan and Blodow, Nico and Marton, Zoltan Csaba and Beetz, Michael},
  booktitle={2008 IEEE/RSJ international conference on intelligent robots and systems},
  pages={3384--3391},
  year={2008},
  organization={IEEE}
}

@article{salti2014shot,
  title={SHOT: Unique signatures of histograms for surface and texture description},
  author={Salti, Samuele and Tombari, Federico and Di Stefano, Luigi},
  journal={Computer Vision and Image Understanding},
  volume={125},
  pages={251--264},
  year={2014},
  publisher={Elsevier}
}

@article{johnson1999using,
  title={Using spin images for efficient object recognition in cluttered 3D scenes},
  author={Johnson, Andrew E and Hebert, Martial},
  journal={IEEE Transactions on pattern analysis and machine intelligence},
  volume={21},
  number={5},
  pages={433--449},
  year={1999},
  publisher={IEEE}
}

@inproceedings{yin2018locnet,
  title={Locnet: Global localization in 3d point clouds for mobile vehicles},
  author={Yin, Huan and Tang, Li and Ding, Xiaqing and Wang, Yue and Xiong, Rong},
  booktitle={2018 IEEE Intelligent Vehicles Symposium (IV)},
  pages={728--733},
  year={2018},
}

@inproceedings{qi2017pointnet,
  title={Pointnet: Deep learning on point sets for 3d classification and segmentation},
  author={Qi, Charles R and Su, Hao and Mo, Kaichun and Guibas, Leonidas J},
  booktitle={Proceedings of the IEEE conference on computer vision and pattern recognition},
  pages={652--660},
  year={2017}
}

@inproceedings{arandjelovic2016netvlad,
  title={NetVLAD: CNN architecture for weakly supervised place recognition},
  author={Arandjelovic, Relja and Gronat, Petr and Torii, Akihiko and Pajdla, Tomas and Sivic, Josef},
  booktitle={Proceedings of the IEEE conference on computer vision and pattern recognition},
  pages={5297--5307},
  year={2016}
}

@inproceedings{uy2018pointnetvlad,
  title={Pointnetvlad: Deep point cloud based retrieval for large-scale place recognition},
  author={Uy, Mikaela Angelina and Lee, Gim Hee},
  booktitle={Proceedings of the IEEE conference on computer vision and pattern recognition},
  pages={4470--4479},
  year={2018}
}

@inproceedings{schaupp2019oreos,
  title={OREOS: Oriented recognition of 3D point clouds in outdoor scenarios},
  author={Schaupp, Lukas and B{\"u}rki, Mathias and Dub{\'e}, Renaud and Siegwart, Roland and Cadena, Cesar},
  booktitle={2019 IEEE/RSJ International Conference on Intelligent Robots and Systems (IROS)},
  year={2019},
}

@article{chen2021overlapnet,
  title={OverlapNet: Loop closing for LiDAR-based SLAM},
  author={Chen, Xieyuanli and L{\"a}be, Thomas and Milioto, Andres and R{\"o}hling, Timo and Vysotska, Olga and Haag, Alexandre and Behley, Jens and Stachniss, Cyrill},
  journal={arXiv preprint arXiv:2105.11344},
  year={2021}
}

@article{xu2021disco,
  title={Disco: Differentiable scan context with orientation},
  author={Xu, Xuecheng and Yin, Huan and Chen, Zexi and Li, Yuehua and Wang, Yue and Xiong, Rong},
  journal={IEEE Robotics and Automation Letters},
  volume={6},
  number={2},
  pages={2791--2798},
  year={2021},
  publisher={IEEE}
}

@article{cramariuc2018learning,
  title={Learning 3D segment descriptors for place recognition},
  author={Cramariuc, Andrei and Dub{\'e}, Renaud and Sommer, Hannes and Siegwart, Roland and Gilitschenski, Igor},
  journal={arXiv preprint arXiv:1804.09270},
  year={2018}
}

@article{dube2020segmap,
  title={SegMap: Segment-based mapping and localization using data-driven descriptors},
  author={Dube, Renaud and Cramariuc, Andrei and Dugas, Daniel and Sommer, Hannes and Dymczyk, Marcin and Nieto, Juan and Siegwart, Roland and Cadena, Cesar},
  journal={The International Journal of Robotics Research},
  volume={39},
  number={2-3},
  pages={339--355},
  year={2020},
  publisher={Sage Publications Sage UK: London, England}
}

@article{kim20191,
  title={1-day learning, 1-year localization: Long-term lidar localization using scan context image},
  author={Kim, Giseop and Park, Byungjae and Kim, Ayoung},
  journal={IEEE Robotics and Automation Letters},
  volume={4},
  number={2},
  pages={1948--1955},
  year={2019},
  publisher={IEEE}
}

@article{kim2021scan,
  title={Scan context++: Structural place recognition robust to rotation and lateral variations in urban environments},
  author={Kim, Giseop and Choi, Sunwook and Kim, Ayoung},
  journal={IEEE Transactions on Robotics},
  year={2021},
  publisher={IEEE}
}

@article{carlevaris2016university,
  title={University of Michigan North Campus long-term vision and lidar dataset},
  author={Carlevaris-Bianco, Nicholas and Ushani, Arash K and Eustice, Ryan M},
  journal={The International Journal of Robotics Research},
  volume={35},
  number={9},
  year={2016},
  publisher={Sage Publications Sage UK: London, England}
}

@inproceedings{kim2020mulran,
  title={Mulran: Multimodal range dataset for urban place recognition},
  author={Kim, Giseop and Park, Yeong Sang and Cho, Younghun and Jeong, Jinyong and Kim, Ayoung},
  booktitle={2020 IEEE International Conference on Robotics and Automation (ICRA)},
  pages={6246--6253},
  year={2020},
  organization={IEEE}
}

@inproceedings{barnes2020oxford,
  title={The oxford radar robotcar dataset: A radar extension to the oxford robotcar dataset},
  author={Barnes, Dan and Gadd, Matthew and Murcutt, Paul and Newman, Paul and Posner, Ingmar},
  booktitle={2020 IEEE International Conference on Robotics and Automation (ICRA)},
  pages={6433--6438},
  year={2020},
  organization={IEEE}
}

@inproceedings{lusha2022full,
  title={Full version with appendix},
  note={\url{https://drive.google.com/file/d/1OeYMAHucGQg8-CmV5H3gqxKmEd1B4UDs/view?usp=sharing}}
}

\end{document}